\documentclass[letterpaper, 10 pt, conference]{ieeeconf}  

\IEEEoverridecommandlockouts                              
\usepackage{amsfonts}
\usepackage{amsmath}
\usepackage{booktabs}
\usepackage{graphicx}
\usepackage{hyperref}
\usepackage{multirow}
\usepackage{tabularx}
\usepackage{adjustbox}

\usepackage{subcaption}
\usepackage{hyperref}
\usepackage{listings}

\usepackage[style=ieee]{biblatex}
\newcommand{\name}{\textbf{OSMa-Bench++}}

\title{\LARGE \bf 
    OSMa-Bench++: Toward Open-Ended Benchmarking of Semantic Mapping for Manipulation with Prompt-Generated Synthetic Scenes
}

\author{Regina Kurkova$^{1*}$, Maxim Popov$^{1}$ and Sergey Kolyubin$^{1}$
\thanks{$^{1}$ Biomechatronics and Energy-Efficient Robotics (BE2R) Lab, ITMO University, Saint Petersburg, Russia}
\thanks{$^{*}$ Corresponding author: rekurkova@itmo.ru}
}

\begin{document}
\maketitle
\thispagestyle{empty}
\pagestyle{empty}


\begin{abstract}

Semantic mapping methods are increasingly used as intermediate scene representations for downstream robotic reasoning and manipulation, yet their evaluation is still largely tied to fixed benchmark datasets with limited coverage of manipulation-relevant corner cases. In this work, we extend OSMa-Bench~\cite{popov2025osmabench} toward controllable benchmarking with prompt-generated synthetic indoor scenes. Our pipeline automatically generates scene descriptions, synthesizes corresponding environments with SceneSmith~\cite{scenesmith2026}, and adapts the resulting assets into an OSMa-Bench-compatible simulation format. This adaptation requires a nontrivial intermediate layer, including semantic normalization, material and texture repair, shader fallback policies, floor handling, navigation setup, and controlled lighting configuration. A key advantage of the proposed setup is that the original scene-generation prompt is known in advance and can therefore serve as an auxiliary semantic specification of the intended scene. We use this property to extend the VQA component of OSMa-Bench with a prompt-grounded question category. The resulting framework supports targeted stress-testing of semantic scene representations under conditions such as clutter, small objects, partial occlusions, and lighting variation, and makes benchmarking more extensible and better aligned with downstream manipulation requirements. Our code is available at \url{https://github.com/be2rlab/OSMa-Bench-v2}.
\end{abstract}

\section{Introduction}

Semantic mapping is increasingly used not only for perception, but also as an intermediate representation for downstream robotic reasoning and manipulation. In this setting, it is not sufficient to produce a visually plausible reconstruction or a semantically labeled map: the representation must preserve object presence, relations, layout, and accessibility cues required for action-oriented queries.

Recent progress in open-vocabulary and object-centric scene representation has substantially improved the expressiveness of robotic maps. Methods such as OpenScene~\cite{peng2023openscene}, OpenMask3D~\cite{takmaz2023openmask3d}, OpenIns3D~\cite{huang2023openins3d}, PLA~\cite{ding2023pla}, ConceptGraphs~\cite{gu2024conceptgraphs}, and BBQ~\cite{linok2024beyond} show that vision-language features, 3D reconstruction, and scene graph abstractions can support flexible semantic querying in complex environments. At the same time, evaluation of such methods remains constrained by fixed datasets and closed scene collections. However, Habitat-based datasets and embodied QA benchmarks still underrepresent manipulation-relevant corner cases such as cluttered small objects, partial occlusions, and restricted access to target objects.

In our previous work on OSMa-Bench~\cite{popov2025osmabench}, we observed that such corner cases can expose substantial differences between methods even when their behavior appears similar on standard scenes. This motivates moving from fixed-scene evaluation toward controllable generation of targeted scenarios. In this work, we extend OSMa-Bench with a prompt-driven synthetic scene generation pipeline based on SceneSmith~\cite{scenesmith2026}. We generate indoor scene descriptions, synthesize corresponding scenes, adapt them to an OSMa-Bench-compatible simulation format, and evaluate semantic representations under multiple conditions.

A key advantage of the proposed setup is that the original scene-generation prompt is known in advance and can therefore serve as an auxiliary semantic specification of the intended scene. We use this property to extend the VQA component of OSMa-Bench with a prompt-grounded category, enabling evaluation not only of what is visible in rendered trajectories, but also of how well the recovered scene representation remains consistent with the intended scene structure. Overall, the proposed framework makes benchmarking more extensible by enabling targeted generation of challenging manipulation-oriented evaluation scenarios.
\section{Methodology}
\label{sec:methodology}

\begin{figure*}[t]
    \centering
    \includegraphics[width=0.98\textwidth]{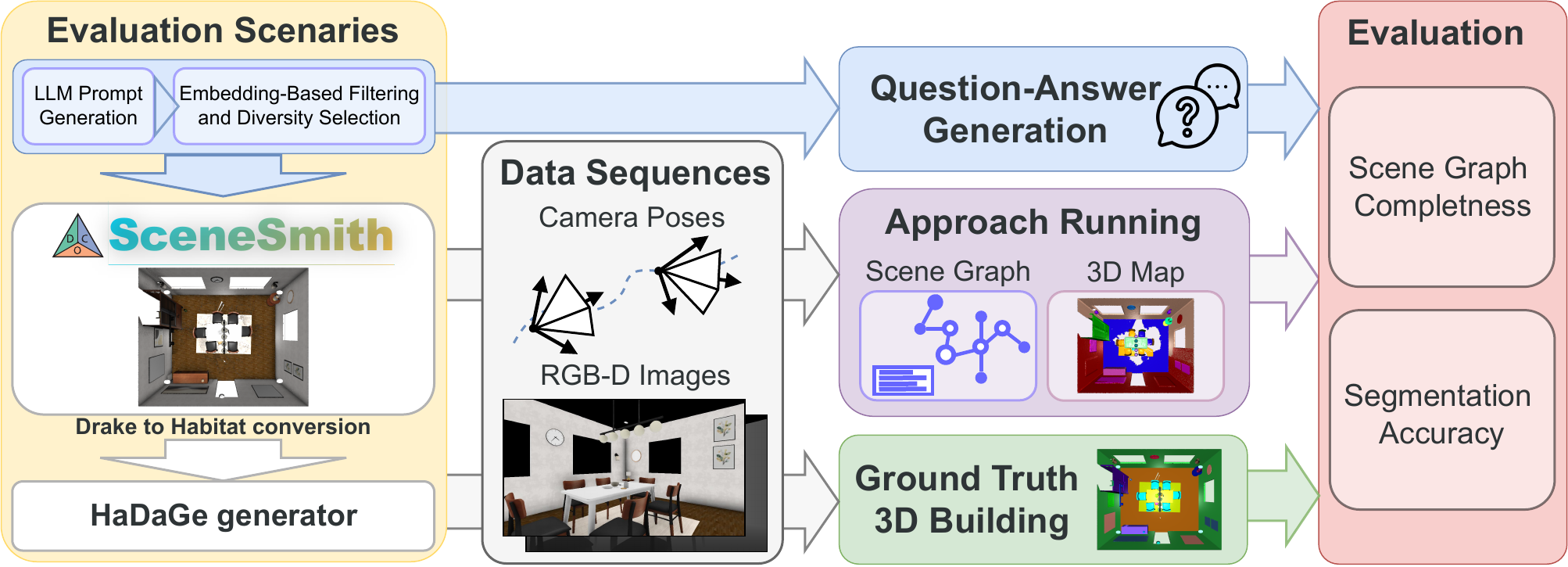}
    \caption{Overview of the \name{} pipeline. Scene descriptions are generated and filtered to obtain diverse evaluation scenarios, synthesized with SceneSmith, converted for use in Habitat-based sequence generation with HaDaGe, processed by semantic mapping methods, and evaluated through segmentation and scene graph completeness metrics.}
    \label{fig:full_pipeline}
\end{figure*}

Our pipeline \name{} extends original method with controllable generation of synthetic indoor scenes for manipulation-oriented evaluation of semantic scene representations. The overall workflow is illustrated in Fig.~\ref{fig:full_pipeline}. Starting from textual scene descriptions, we synthesize indoor environments with SceneSmith~\cite{scenesmith2026}, convert the resulting assets into a Habitat-compatible format, generate RGB-D observation sequences with the HaDaGe generator~\cite{popov2025osmabench} used in OSMa-Bench, run semantic mapping approaches on these sequences, and finally evaluate the resulting representations through segmentation metrics and VQA-based scene graph assessment.

We begin from automatic generation of scene descriptions, corresponding to the prompt generation block in Fig.~\ref{fig:full_pipeline}. A large language model first produces an initial pool of candidate prompts describing indoor scenes with object layouts and relations relevant to downstream manipulation. These prompts are then embedded into a semantic feature space and compared using cosine similarity,
\begin{equation}
\mathrm{sim}(p_i,p_j)=\frac{f(p_i)^\top f(p_j)}{\|f(p_i)\|_2 \, \|f(p_j)\|_2},
\end{equation}
where $f(\cdot)$ denotes the text embedding model.

This representation is used in two stages. First, near-duplicate prompts are removed to avoid repeatedly generating semantically similar scenes. Second, from the remaining pool we select the most diverse subset by favoring prompts that are far from those already chosen in the embedding space. In practice, this yields a set of scene descriptions that covers a broader range of layouts and interaction-relevant situations than random sampling or manual prompt writing alone.

Since SceneSmith outputs are in Drake~\cite{tedrake2019drake} simulation format and are not directly compatible with OSMa-Bench, we convert them into a Habitat-like dataset structure, constructing scene and object configuration files, assigning semantic categories, and resolving asset-level inconsistencies. After conversion, the scenes are passed to the OSMa-Bench~\cite{popov2025osmabench}, which produces RGB-D observation sequences along camera trajectories.

Finally, the generated sequences are processed by the evaluated approaches, which produce 3D maps and scene graphs. The benchmark evaluates both segmentation quality and scene graph completeness through question answering. 

In addition to standard VQA categories, we use the original scene-generation prompt as an auxiliary source of semantic specification for question generation. This makes it possible to probe not only what is visible in the rendered sequence, but also how well the recovered representation remains consistent with the intended scene layout described at the prompt level.
\section{Experiments}

\subsection{Experimental Setup}

We evaluate the proposed extension on SceneSmith-generated indoor scenes integrated into OSMa-Bench. We consider two open semantic mapping methods, ConceptGraphs~\cite{gu2024conceptgraphs} and BBQ~\cite{linok2024beyond}, which produce structured scene representations for downstream question answering.

The evaluation set is constructed from automatically generated prompts.From an initial pool of 350 prompts, we retain diverse candidates using the filtering procedure in Section~\ref{sec:methodology} and keep only technically valid scenes. The final benchmark contains 40 scenes, split into 24 \emph{furniture} scenes and 16 \emph{manipuland} scenes.
Prompt-to-scene correspondence was manually checked for both categories. Scenes with substantial mismatches were excluded during quality control. In the final set of 40 scenes, only four minor synthesis errors were observed, each corresponding to one missing object. This gives a scene-level full-match rate of $36/40 = 90.0\%$. Assuming at least five prompted objects per retained scene, this corresponds to a conservative lower-bound object-level fidelity of at least $98.0\%$.

We generate two complementary scene families. The \emph{furniture} subset is used to evaluate whether a representation preserves room-scale structure, object layout, support relations, and coarse spatial dependencies in scenes dominated by large static objects. This subset is also useful early in synthesis because such scenes are easier to generate and export. The \emph{manipuland} subset adds small interaction-relevant objects placed on tables, shelves, counters, and other supporting surfaces. These scenes are more challenging because they increase clutter, introduce fine-grained support and proximity relations, and include small, partially occluded, or densely arranged objects. Together, the two subsets separate room-scale structural failures from small-object and manipulation-oriented failures. For prompt generation and prompt-grounded question generation, we use GPT-4.1~\cite{openai2025gpt41}.

Following OSMa-Bench~\cite{popov2025osmabench}, we evaluate each scene under four lighting conditions. The \textit{baseline} uses static, non-uniformly distributed light sources natively available in the scene. The \textit{nominal lights} configuration removes all explicit sources, relying solely on mesh-emitted light. The \textit{camera light} condition attaches a directed light source to the camera, producing strong view-dependent shadows. Finally, \textit{dynamic lighting} varies illumination along the robot trajectory, simulating realistic transitions between differently lit areas.

\begin{figure}[t]
    \centering
    \includegraphics[width=0.49\textwidth]{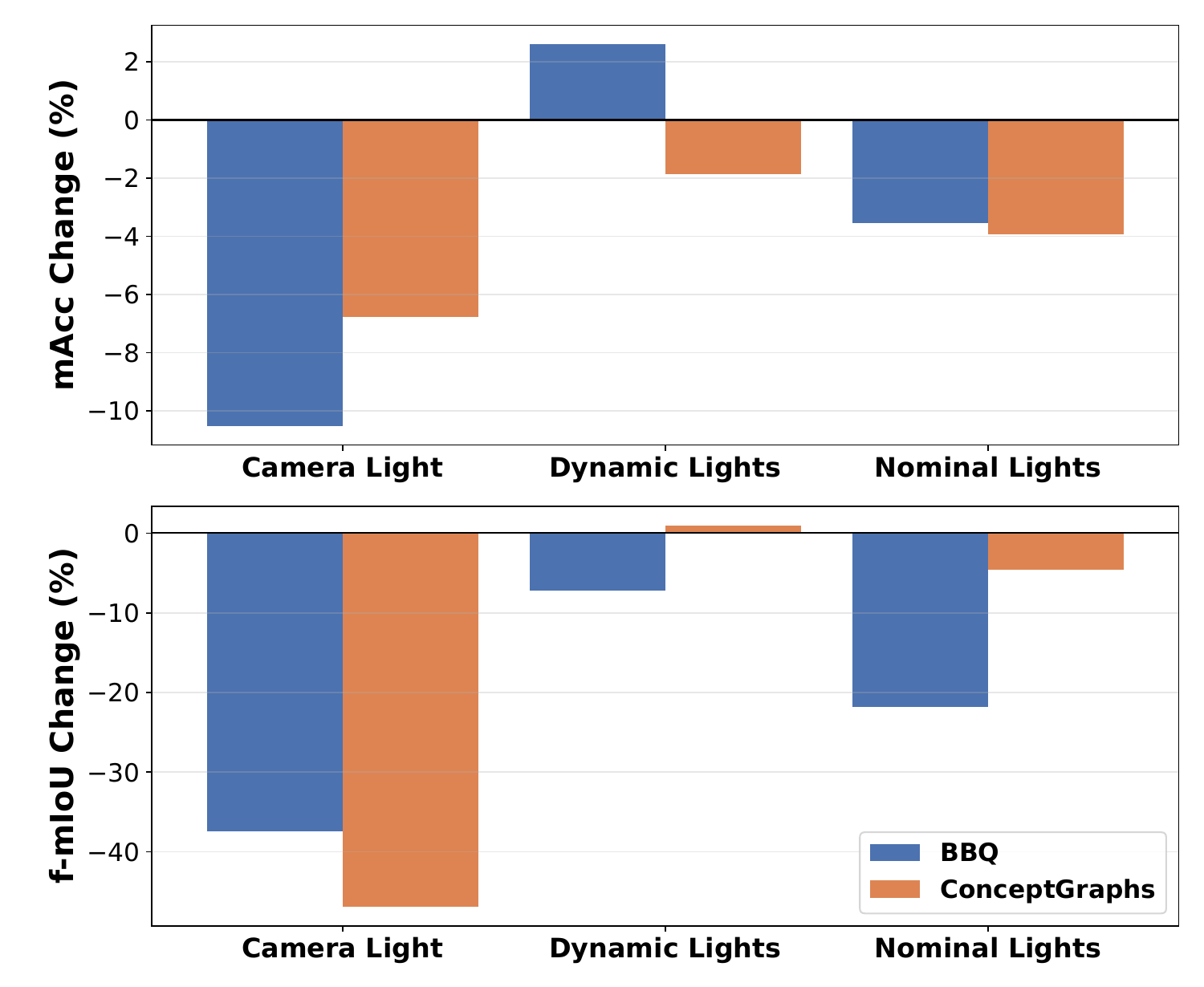}
    \caption{Illustrative results on SceneSmith-derived \name{} scenes. Top: relative change in semantic segmentation performance (mAcc) for BBQ~\cite{linok2024beyond} and ConceptGraphs~\cite{gu2024conceptgraphs} under different lighting conditions with respect to the baseline. Bottom: corresponding change in f-mIoU.}
    \label{fig:semseg}
\end{figure}

\subsection{Prompt-Grounded Evaluation Protocol}

In addition to the standard OSMa-Bench evaluation, we introduce a prompt-grounded protocol in which the original scene description serves as an auxiliary source of semantic ground truth. This protocol is particularly useful for categories that are difficult to assess robustly in standard trajectory-based VQA. In the original OSMa-Bench setting, the answer to a question may depend on what is visible from a specific trajectory or viewpoint. This is especially problematic for questions about object counts and inter-object relations, because the relevant objects may be present in the scene but only partially visible, occluded, or never jointly observable from the chosen camera path. As a result, these categories are among the least stable ones when questions are generated purely from view-dependent observations.

For this reason, we focus the prompt-grounded analysis on the \emph{Measurements} and \emph{Relations} between objects categories. These two categories are the most informative for scene graph completeness and are also the most strongly affected by the limitations of view-dependent question generation. Measurement questions probe whether the representation preserves the number of relevant entities, which is sensitive to missed small objects, duplicated instances, and incorrect object merging. Relation questions probe whether the representation preserves the spatial and structural dependencies between entities, such as support, proximity, containment, and relative arrangement. In a prompt-grounded setting, these questions no longer depend on whether a specific camera trajectory happened to expose the required relation.

This design is especially important for manipulation-oriented evaluation. For downstream manipulation, it is often not enough to know that an object exists; the representation must also preserve how many relevant objects are present and how they are arranged with respect to one another. The prompt-grounded protocol therefore complements the original OSMa-Bench VQA evaluation by targeting categories that are semantically central yet difficult to assess robustly through view-dependent observations alone.

\subsection{Results on Segmentation and Prompt-Grounded Evaluation}

Table~\ref{tab:semseg} and Fig.~\ref{fig:semseg} show the relative change in segmentation performance with respect to \emph{Baseline}. \emph{Camera} lighting produces the strongest degradation for both methods, while \emph{Dynamic} lighting has a mild effect with small positive changes in mAcc. The divergence between mAcc and f-mIoU reflects a structural difference between the two approaches: BBQ tends to produce fewer but geometrically precise masks, yielding high f-mIoU yet lower scene-graph completeness, since some objects are filtered out entirely. ConceptGraphs recovers more objects overall, resulting in higher completeness but less precise masks. These complementary failure modes confirm that segmentation robustness cannot be assessed from a single metric alone.

Table~\ref{tab:promptgt_accuracy_grouped} reports prompt-grounded question answering accuracy for BBQ and ConceptGraphs on the \emph{Measurements} and \emph{Relations} categories, separated into the \emph{Furniture} and \emph{Manipuland} subsets. Overall, both methods score below 30\%, yet reveal distinct robustness profiles. BBQ, being a more advanced scene-graph method with stronger question-answering capabilities, generally outperforms ConceptGraphs on \emph{Relations} and on the \emph{Manipuland} subset. However, BBQ shows a notable drop under \emph{Dynamic} lighting, which we attribute to its object description strategy: BBQ generates per-object embeddings from a single image, so a frame captured under unfavorable conditions directly degrades the object's representation. ConceptGraphs, by contrast, aggregates evidence across multiple views, which stabilizes its embeddings under \emph{Dynamic} lighting and explains its gains in that condition, particularly on \emph{Furniture} categories.

Table~\ref{tab:question_source_ablation} compares standard image-derived VQA from OSMa-bench with prompt-grounded questions for the same categories. Unlike standard VQA, which inherits trajectory-dependent visibility and perception biases, PromptGT directly evaluates prompt-specified counts and relations.

\begin{table*}[t]
    \centering
    \small
    \setlength{\tabcolsep}{5pt}
    \caption{Semantic segmentation performance on generated sequences under different lighting conditions.}
    \label{tab:semseg}
    \begin{tabular}{@{}l|cc|cc|cc|cc@{}}
        \toprule
        \multirow{2}{*}{\textbf{Method}} & 
        \multicolumn{2}{c|}{\textbf{Baseline}} & 
        \multicolumn{2}{c|}{\textbf{Camera Light}} & 
        \multicolumn{2}{c|}{\textbf{Dynamic Lights}} & 
        \multicolumn{2}{c}{\textbf{Nominal Lights}} \\
        \cmidrule(lr){2-3} \cmidrule(lr){4-5} \cmidrule(lr){6-7} \cmidrule(lr){8-9}
        & \textbf{mAcc} & \textbf{f-mIoU} & \textbf{mAcc} & \textbf{f-mIoU} & \textbf{mAcc} & \textbf{f-mIoU} & \textbf{mAcc} & \textbf{f-mIoU} \\
        \midrule
        ConceptGraphs~\cite{gu2024conceptgraphs}
        & \textbf{57.6} & 27.9
        & \textbf{53.7} & 14.8
        & \textbf{56.5} & 28.0
        & \textbf{55.3} & 26.6 \\
        BBQ~\cite{linok2024beyond}
        & 47.0 & \textbf{68.4}
        & 42.0 & \textbf{42.8}
        & 48.2 & \textbf{63.5}
        & 45.3 & \textbf{53.5} \\
        \bottomrule
    \end{tabular}
\end{table*}
\begin{table}[t]
    \centering
    \caption{Accuracy (\%) of prompt-grounded question answering for BBQ~\cite{linok2024beyond} and ConceptGraphs~\cite{gu2024conceptgraphs}. Values are computed over all questions in each category across 40 scenes.}
    \label{tab:promptgt_accuracy_grouped}
    \setlength{\tabcolsep}{4pt}
    \begin{tabular}{@{}l|c|c|cccc@{}}
        \toprule
        \textbf{Subset} & \textbf{Cat.} & \textbf{Method} & \textbf{Baseline} & \textbf{Camera} & \textbf{Dynamic} & \textbf{Nominal} \\
        \midrule
        \multirow{4}{*}{\textbf{Furn.}} & \multirow{2}{*}{Measur.} & CG & 12.2 & 9.2 & \textbf{15.3} & \textbf{14.3} \\
         &  & BBQ & \textbf{15.3} & \textbf{24.5} & 12.2 & 12.2 \\
        \cmidrule(lr){2-2} \cmidrule(lr){3-3} \cmidrule(l){4-7}
         & \multirow{2}{*}{Relat.} & CG & 12.9 & 6.4 & \textbf{20.5} & 17.0 \\
         &  & BBQ & \textbf{18.7} & \textbf{24.6} & 14.6 & \textbf{18.1} \\
        \midrule
        \multirow{4}{*}{\textbf{Manip.}} & \multirow{2}{*}{Measur.} & CG & 12.9 & 11.4 & 17.1 & 12.9 \\
         &  & BBQ & \textbf{15.7} & \textbf{24.3} & \textbf{18.6} & \textbf{22.9} \\
        \cmidrule(lr){2-2} \cmidrule(lr){3-3} \cmidrule(l){4-7}
         & \multirow{2}{*}{Relat.} & CG & 15.0 & 14.0 & 13.0 & 9.0 \\
         &  & BBQ & \textbf{16.0} & \textbf{21.0} & \textbf{14.0} & \textbf{19.0} \\
        \bottomrule
    \end{tabular}
\end{table}
\begin{table}[t]
    \centering
    \caption{Question-source ablation for measurement and relational QA Accuracy. Standard denotes image-derived VQA from OSMa-Bench, while PromptGT denotes questions generated from the scene-generation prompt.}
    \label{tab:question_source_ablation}
    \setlength{\tabcolsep}{5pt}
    \begin{tabular}{@{}l|c|c|cc@{}}
        \toprule
        \textbf{Subset} & \textbf{Cat.} & \textbf{Method} & \textbf{Standard} & \textbf{PromptGT} \\
        \midrule
        \multirow{4}{*}{\textbf{Furn.}} 
        & \multirow{2}{*}{Measur.} & CG  & \textbf{30.6} & 12.2 \\
        &                          & BBQ & \textbf{18.4 }& 15.3 \\
        \cmidrule(lr){2-2} \cmidrule(lr){3-3} \cmidrule(l){4-5}
        & \multirow{2}{*}{Relat.}  & CG  & \textbf{19.7} & 12.9 \\
        &                          & BBQ & \textbf{24.6} & 18.7 \\
        \midrule
        \multirow{4}{*}{\textbf{Manip.}} 
        & \multirow{2}{*}{Measur.} & CG  & \textbf{32.2} & 12.9 \\
        &                          & BBQ & \textbf{33.9} & 15.7 \\
        \cmidrule(lr){2-2} \cmidrule(lr){3-3} \cmidrule(l){4-5}
        & \multirow{2}{*}{Relat.}  & CG  & \textbf{22.2} & 15.0 \\
        &                          & BBQ & \textbf{26.3} & 16.0 \\
        \bottomrule
    \end{tabular}
\end{table}

Overall, these experiments highlight the main advantage of the proposed extension. By generating scenes from prompts, we produce more diverse evaluation environments than a fixed benchmark and assess categories that are difficult to evaluate in a view-dependent setup. In particular, prompt-grounded \emph{Measurements} and \emph{Relations} between objects provide a more stable probe of scene graph completeness, while the relative-to-baseline semantic segmentation analysis shows how these graph-level effects relate to the quality of the underlying 3D semantic reconstruction.
\section{Conclusion}
We presented \name{}, an extension of original OSMa-Bench that enables controllable benchmarking of semantic mapping methods with prompt-generated synthetic indoor scenes. By combining prompt generation, SceneSmith-based scene synthesis, Habitat-compatible conversion, and HaDaGe-based sequence generation, the proposed pipeline makes it possible to construct evaluation scenarios that are difficult to obtain in fixed datasets, including cluttered layouts, small interaction-relevant objects, and relation-heavy scenes. A central property of this setup is that the original scene-generation prompt is known a priori and can therefore be reused as an auxiliary semantic specification for evaluation.

The experiments show that this addition is useful in two ways. First, it expands the benchmark beyond a closed set of scenes and makes targeted stress-testing possible under controlled conditions. Second, the prompt-grounded protocol provides a more stable way to assess scene graph completeness for measurements and inter-object relations, which are otherwise strongly affected by trajectory-dependent visibility. The results also indicate that BBQ and ConceptGraphs respond differently to lighting changes and scene composition, confirming that graph-level semantic completeness and segmentation robustness capture complementary aspects of representation quality. Overall, \textbf{OSMa-Bench++} turns synthetic scene generation into a practical tool for manipulation-oriented evaluation of semantic scene representations.



\printbibliography

@inproceedings{gu2024conceptgraphs,
  title={Conceptgraphs: Open-vocabulary 3d scene graphs for perception and planning},
  author={Gu, Qiao and Kuwajerwala, Ali and Morin, Sacha and Jatavallabhula, Krishna Murthy and Sen, Bipasha and Agarwal, Aditya and Rivera, Corban and Paul, William and Ellis, Kirsty and Chellappa, Rama and others},
  booktitle={2024 IEEE International Conference on Robotics and Automation (ICRA)},
  pages={5021--5028},
  year={2024},
  organization={IEEE}
}

@misc{tedrake2019drake,
  title={Drake: Model-based design and verification for robotics},
  author={Tedrake, Russ and others},
  year={2019}
}

@techreport{openai2025gpt41,
  author       = {OpenAI},
  title        = {Introducing GPT‑4.1 in the API},
  institution  = {OpenAI},
  year         = {2025},
  url          = {https://openai.com/index/gpt-4-1/},
}

@inproceedings{popov2025osmabench,
    title={OSMa-Bench: Evaluating Open Semantic Mapping Under Varying Lighting Conditions},
    author={Popov, Maxim and Kurkova, Regina and Iumanov, Mikhail and Mahmoud, Jaafar and Kolyubin, Sergey},
    booktitle={2025 IEEE/RSJ International Conference on Intelligent Robots and Systems (IROS)},
    year={2025}
}

@misc{scenesmith2026,
  title={SceneSmith: Agentic Generation of Simulation-Ready Indoor Scenes},
  author={Nicholas Pfaff and Thomas Cohn and Sergey Zakharov and Rick Cory and Russ Tedrake},
  year={2026},
  eprint={2602.09153},
  archivePrefix={arXiv},
  primaryClass={cs.RO},
  url={https://arxiv.org/abs/2602.09153},
}

@article{takmaz2023openmask3d,
  title={Openmask3d: Open-vocabulary 3d instance segmentation},
  author={Takmaz, Ay{\c{c}}a and Fedele, Elisabetta and Sumner, Robert W and Pollefeys, Marc and Tombari, Federico and Engelmann, Francis},
  journal={arXiv preprint arXiv:2306.13631},
  year={2023}
}

@inproceedings{ding2023pla,
  title={Pla: Language-driven open-vocabulary 3d scene understanding},
  author={Ding, Runyu and Yang, Jihan and Xue, Chuhui and Zhang, Wenqing and Bai, Song and Qi, Xiaojuan},
  booktitle={Proceedings of the IEEE/CVF conference on computer vision and pattern recognition},
  pages={7010--7019},
  year={2023}
}

@article{huang2023openins3d,
  title={Openins3d: Snap and lookup for 3d open-vocabulary instance segmentation},
  author={Huang, Zhening and Wu, Xiaoyang and Chen, Xi and Zhao, Hengshuang and Zhu, Lei and Lasenby, Joan},
  journal={arXiv preprint arXiv:2309.00616},
  year={2023}
}

@inproceedings{peng2023openscene,
  title={Openscene: 3d scene understanding with open vocabularies},
  author={Peng, Songyou and Genova, Kyle and Jiang, Chiyu and Tagliasacchi, Andrea and Pollefeys, Marc and Funkhouser, Thomas and others},
  booktitle={Proceedings of the IEEE/CVF conference on computer vision and pattern recognition},
  pages={815--824},
  year={2023}
}

@article{linok2024beyond,
  title={Beyond Bare Queries: Open-Vocabulary Object Grounding with 3D Scene Graph},
  author={Linok, Sergey and Zemskova, Tatiana and Ladanova, Svetlana and Titkov, Roman and Yudin, Dmitry and Monastyrny, Maxim and Valenkov, Aleksei},
  journal={arXiv preprint arXiv:2406.07113},
  year={2024}
}






\end{document}